\documentclass{article}
\usepackage{authblk}
\usepackage[utf8]{inputenc}
\usepackage{main}
\usepackage{microtype}
\usepackage{subcaption}
\usepackage{graphicx}
\usepackage{times}
\usepackage{latexsym}
\usepackage{amsmath}
\usepackage{float}
\usepackage{footnote}
\usepackage{enumitem}
\usepackage{bm}
\usepackage{arydshln}
\usepackage{booktabs}
\usepackage{multicol}
\usepackage{multirow}
\usepackage{color}
\usepackage{xcolor}     
\usepackage{colortbl}
\usepackage{makecell}
\usepackage{mathtools}
\usepackage{imakeidx}
\usepackage{longtable}
\usepackage{wrapfig}
\usepackage{rotating}
\makeindex
\usepackage{arydshln}
\usepackage{lipsum}
\usepackage{natbib}
\usepackage{xspace}
\newcommand{\method}{UMoE\xspace}
\newcommand{\directsft}{Direct SFT\xspace}
\usepackage[toc]{multitoc}
\usepackage[edges]{forest}
\usepackage[normalem]{ulem}
\definecolor{mydarkblue}{rgb}{0,0.08,0.45}
\usepackage[colorlinks=true,linkcolor=mydarkblue,citecolor=mydarkblue,filecolor=mydarkblue,urlcolor=mydarkblue]{hyperref}
\usepackage[most]{tcolorbox}
\usepackage{booktabs}
\usepackage{geometry}
\definecolor{wkblue}{rgb}{0.2, 0.3, 0.6}
\definecolor{meta-color}{rgb}{0.5, 0.5, 0.5}
\usepackage{amsmath}
\usepackage{enumitem}
\usepackage{lscape} 
\usepackage{booktabs}
\newfloat{algorithm}{t}{loa}
\floatname{algorithm}{Algorithm}
\usepackage{pifont}
\usepackage{tabularx,booktabs}
\usepackage{makecell}

\usepackage{amssymb}
\usepackage{amsfonts}

\usepackage[framemethod=tikz]{mdframed}
\definecolor{bgblue}{RGB}{245,243,253}
\definecolor{ttblue}{RGB}{91,194,224}
\definecolor{codegreen}{rgb}{0,0.6,0}
\definecolor{codegray}{rgb}{0.5,0.5,0.5}
\definecolor{codepurple}{rgb}{0.58,0,0.82}
\definecolor{backcolour}{rgb}{0.95,0.95,0.92}
\definecolor{wkgreen}{RGB}{184,244,175}
\definecolor{wkpurple}{RGB}{210,210,253}
\definecolor{wkyellow}{RGB}{255,241,177}
\definecolor{wkblue}{RGB}{174,217,253}
\definecolor{mydarkblue}{rgb}{0,0.08,0.45}
\definecolor{wkblue}{rgb}{0.2, 0.3, 0.6}
\definecolor{dpblue}{rgb}{0.0, 0.333, 0.643}

\mdfdefinestyle{mystyle}{%
  rightline=true,
  innerleftmargin=10,
  innerrightmargin=10,
  outerlinewidth=3pt,
  topline=false,
  rightline=true,
  bottomline=false,
  skipabove=\topsep,
  skipbelow=\topsep
}

\newtcolorbox{myboxi}[1][]{
  breakable,
  title=#1,
  colback=red!5,
  colbacktitle=red!5,
  coltitle=black,
  fonttitle=\bfseries,
  bottomrule=0pt,
  toprule=0pt,
  leftrule=2pt,
  rightrule=2pt,
  titlerule=0pt,
  arc=0pt,
  outer arc=0pt,
  colframe=red,
}

\newtcolorbox{myboxnote}[1][]{
  breakable,
  title=#1,
  colback=orange!0,
  colbacktitle=orange!0,
  coltitle=black,
  fonttitle=\bfseries,
  bottomrule=0pt,
  toprule=0pt,
  leftrule=2pt,
  rightrule=2pt,
  titlerule=0pt,
  arc=0pt,
  outer arc=0pt,
  colframe=orange,
}

\newtcolorbox{myboxii}[1][]{
  breakable,
  freelance,
  title=#1,
  colback=white,
  colbacktitle=white,
  coltitle=black,
  fonttitle=\bfseries,
  bottomrule=0pt,
  boxrule=0pt,
  colframe=white,
  overlay unbroken and first={
  \draw[red!75!black,line width=3pt]
    ([xshift=5pt]frame.north west) -- 
    (frame.north west) -- 
    (frame.south west);
  \draw[red!75!black,line width=3pt]
    ([xshift=-5pt]frame.north east) -- 
    (frame.north east) -- 
    (frame.south east);
  },
  overlay unbroken app={
  \draw[red!75!black,line width=3pt,line cap=rect]
    (frame.south west) -- 
    ([xshift=5pt]frame.south west);
  \draw[red!75!black,line width=3pt,line cap=rect]
    (frame.south east) -- 
    ([xshift=-5pt]frame.south east);
  },
  overlay middle and last={
  \draw[red!75!black,line width=3pt]
    (frame.north west) -- 
    (frame.south west);
  \draw[red!75!black,line width=3pt]
    (frame.north east) -- 
    (frame.south east);
  },
  overlay last app={
  \draw[red!75!black,line width=3pt,line cap=rect]
    (frame.south west) --
    ([xshift=5pt]frame.south west);
  \draw[red!75!black,line width=3pt,line cap=rect]
    (frame.south east) --
    ([xshift=-5pt]frame.south east);
  },
}

\usepackage{fancyhdr} 
\usepackage{blindtext} 
\usepackage{makecell}

\DeclareCaptionFont{black}{\color{black}}

\definecolor{myblue}{rgb}{0.9, 0.1, 0.94}
\definecolor{mygreen}{rgb}{0.64, 0.56, 0.88}
\definecolor{myyellow}{rgb}{0.68, 0.6, 0.1}
\definecolor{fancygreen}{rgb}{0.33, 0.68, 0.20}
\definecolor{salmon}{rgb}{0.94, 0.52, 0.49}
\definecolor{tablegreen}{rgb}{0.82, 0.94, 0.75}
\definecolor{tableblue}{rgb}{0.81, 0.90, 0.94}
\definecolor{tablered}{rgb}{0.97, 0.85, 0.85}
\definecolor{tableorange}{rgb}{0.96, 0.85, 0.81}
\definecolor{tir}{rgb}{0.592,0.741,0.988}
\definecolor{cot}{rgb}{0.965,0.443,0.537}
\definecolor{grey}{rgb}{0.502,0.502,0.502}
\definecolor{grey}{rgb}{0.502,0.502,0.502}
\definecolor{darkyellow}{rgb}{0.855,0.647,0.125}

\newenvironment{itemize*}%
 {\leftmargini=10pt\begin{itemize}%
  \setlength{\itemsep}{0pt}%
  \setlength{\parskip}{0pt}%
  }%
 {\end{itemize}}
\newenvironment{enumerate*}%
 {\begin{enumerate}%
  \setlength{\itemsep}{0pt}%
  \setlength{\parskip}{0pt}}%
 {\end{enumerate}}

\usepackage{listings}

\newcommand\JSONnumbervaluestyle{\color{blue}}
\newcommand\JSONstringvaluestyle{\color{red}}

\newif\ifcolonfoundonthisline

\makeatletter

\lstdefinestyle{json}
{
  showstringspaces    = false,
  keywords            = {false,true},
  alsoletter          = 0123456789.,
  morestring          = [s]{"}{"},
  stringstyle         = \ifcolonfoundonthisline\JSONstringvaluestyle\fi,
  MoreSelectCharTable =%
    \lst@DefSaveDef{`:}\colon@json{\processColon@json},
  basicstyle          = \ttfamily,
  keywordstyle        = \ttfamily\bfseries,
}

\newcommand\processColon@json{%
  \colon@json%
  \ifnum\lst@mode=\lst@Pmode%
    \global\colonfoundonthislinetrue%
  \fi
}

\lst@AddToHook{Output}{%
  \ifcolonfoundonthisline%
    \ifnum\lst@mode=\lst@Pmode%
      \def\lst@thestyle{\JSONnumbervaluestyle}%
    \fi
  \fi
  \lsthk@DetectKeywords%
}

\lst@AddToHook{EOL}%
  {\global\colonfoundonthislinefalse}

\makeatother

\usepackage{etoolbox}
\usepackage{natbib}
\usepackage{url}
\newcounter{bibcount}
\makeatletter
\patchcmd{\@lbibitem}{\item[}{\item[\hfil\stepcounter{bibcount}{[\thebibcount]}}{}{}
\renewcommand\NAT@bibsetup%
  [1]{\setlength{\leftmargin}{\bibhang}\setlength{\itemindent}{-\parindent}%
      \setlength{\itemsep}{\bibsep}\setlength{\parsep}{\z@}}
\makeatother

\definecolor{mybrown}{RGB}{128,64,0}

\definecolor{titlecolor}{HTML}{4c9cff}

\begin{document}



\title{Unlocking Every Expert in Domain-Specific Training}

\renewcommand\Authand{\qquad}
\author[1,2,3]{Xuefeng Li}
\author[1,2,3]{Pengfei Liu\textsuperscript{†}}
\affil{SII
\quad \textsuperscript{2}SJTU
\quad \textsuperscript{3}GAIR}
  
\maketitle
\begingroup
\renewcommand{\thefootnote}{†}
\footnotetext{Corresponding author.}
\endgroup
\thispagestyle{fancy}
\fancyhead{}
\lhead{\includegraphics[height=0.7cm]{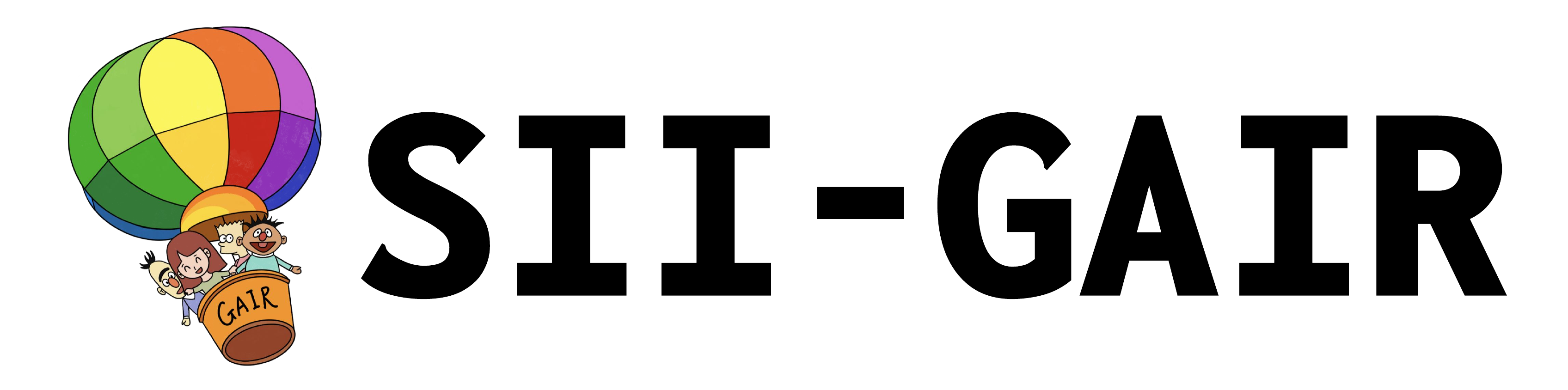}}
\renewcommand{\headrulewidth}{0pt}
\setlength{\headsep}{0mm}

\begin{abstract}
Mixture-of-Experts (MoE) models scale capacity without proportional compute cost and have become a key architecture for frontier large language models (LLMs). Yet domain-specific post-training inherits an expert pool shaped by mixed-domain pre-training: a substantial subset of experts contributes little on the target domain, and standard supervised fine-tuning (SFT) leaves the composition of this pool unchanged.
We propose a simple, budget-preserving pipeline that realigns the expert pool to the target domain before fine-tuning. Given a target domain, we (1) prune the experts with lowest domain-aligned saliency, (2) regrow the expert pool to its original size through perturbation-based expert expansion, and (3) apply standard SFT. The resulting model preserves the original expert count, parameter count, and inference cost.
With a single frozen recipe and no per-domain hyperparameter tuning, \method consistently improves over \directsft across two MoE architectures (Qwen3-30B-A3B and Qwen3.5-35B-A3B), five domains (math, code, science, tool-use, and agentic coding), and 12 benchmarks. Representative improvements are 3.4 points in math average accuracy, 6.0 points on SWE-bench Verified. On a strong in-house math corpus, \directsft already surpasses Qwen3-30B-A3B-Thinking (82.81 vs.\ 81.06), yet \method further raises the average to 84.17, an additional 1.36 points, demonstrating robustness to a substantially stronger SFT regime. Data-scaling experiments further show that the gain persists as training data grows.
Analysis reveals that the direct-SFT model allocates substantial routed-expert compute to a low-saliency subset that can be removed post hoc with little average degradation; \method turns this redundant capacity into useful domain capacity and achieves lower training loss, with gains spanning all difficulty levels in downstream evaluation.
\begin{center}
    \includegraphics[width=0.92\linewidth]{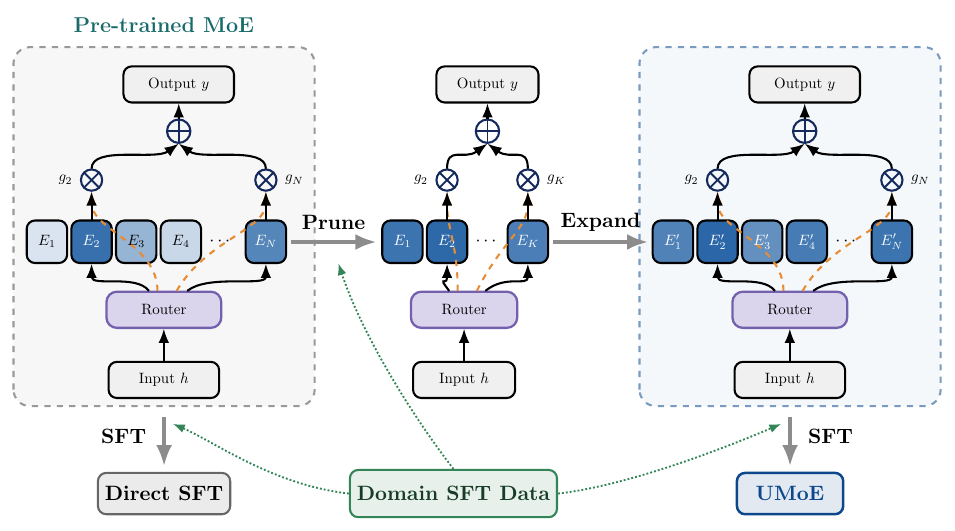}
    \captionsetup{font=footnotesize}
    \captionof{figure}{\textbf{Overview of \method.} Direct SFT fine-tunes the pretrained MoE as is. \method first \textbf{prunes} the least domain-salient experts and \textbf{regrows} the pool from the retained experts, restoring the original size and inference cost before SFT. The same domain data $\mathcal{D}$ is used for calibration and training.}
    \label{fig:teaser}
\end{center}
\end{abstract}

\clearpage

\pagestyle{fancy}
\lhead{\rightmark}
\renewcommand{\headrulewidth}{0.7pt}
\setlength{\headsep}{5mm}

\section{Introduction}
\label{sec:intro}

Frontier large language models such as Qwen3.5~\citep{qwen35}, DeepSeek-V4~\citep{deepseekv4}, and Kimi-K2~\citep{kimi-k2} increasingly adopt Mixture-of-Experts (MoE) architectures~\citep{shazeer2017moe} to expand model capacity while keeping inference efficient. Building on these models, domain-specific post-training serves two important purposes. It can directly push capability on the hardest problems within a target domain~\citep{deepseekmathv2,qwen3coder}; it can also produce specialist models whose capabilities are later consolidated into a generalist through on-policy distillation~\citep{opd,mopd2026} or rejection-sampling fine-tuning~\citep{qwen3,deepseekv32,mimov2flash,stepfun35,deepseekv4}. Both settings require adapting a general-purpose MoE to a specific target domain.

The expert pool inherited from mixed-domain pre-training, however, is not homogeneous. Sparse routing over diverse data encourages experts to develop different functional specializations~\citep{deepseekmoe,olmoe}, making their relevance uneven under a single target distribution. Consequently, a low-saliency subset can continue to receive non-trivial routing on domain data. Standard SFT updates these inherited experts in place but does not explicitly reallocate expert slots, leaving part of the fixed capacity poorly aligned with the target domain.

We therefore ask: can the expert pool be reorganized into a more domain-aligned initialization before SFT, so that more expert slots become useful for learning target-domain capabilities? We propose \textbf{\method} (\textbf{U}nlocking \textbf{MoE} Capacity), a domain-aware expert-pool reorganization method. \method first uses target-domain calibration data to prune the experts with the lowest domain-aligned saliency~\citep{reap2025,notallexperts}. It then regrows the pool through perturbation-based expert expansion, in the spirit of expert upcycling~\citep{sparseupcycling,upcyclingllm,dropupcycling}, restoring the original expert count before applying standard SFT. The resulting model preserves the original architecture, parameter count, and inference cost.

We evaluate \method across five target domains and multiple training-data scales on Qwen3-30B-A3B, and further validate architecture transfer on Qwen3.5-35B-A3B in math. \method consistently improves benchmark averages across every evaluated domain and architecture, while retaining positive gains at all evaluated data scales. A detailed math-domain analysis further shows that the direct-SFT baseline allocates substantial routed-expert compute to a subset that is largely redundant under post-hoc pruning, whereas \method makes a larger fraction of the expert pool consequential to downstream performance.

\section{Related Work}
\label{sec:related}

\paragraph{MoE compression and expert pruning.} Sparse MoE models typically keep the full expert pool in memory while activating only a small subset of experts for each token. Expert utilization is often highly skewed, and experts can become functionally specialized~\citep{deepseekmoe,olmoe}; consequently, their measured importance can vary substantially under a fixed target distribution. Prior work has explored compressing the expert pool to reduce memory footprint and improve deployment efficiency, either by merging multiple experts into fewer replacements or by pruning experts with low estimated importance~\citep{liu2025moesurvey,notallexperts}. For example, REAP combines router gate values with activation norms to score experts and reports near-lossless results in several evaluated generative settings under $50\%$ expert pruning~\citep{reap2025}. This structure is especially relevant to domain- or task-specific deployment, where the target distribution may rely primarily on a subset of the available experts. Existing methods therefore prune MoEs for particular tasks, languages, or domains: \citet{chen2022taskprune} prune task-specific expert structure, \citet{koishekenov2023langprune} compress multilingual MoEs by removing experts with low relevance to target languages, and \citet{dong2025domainprune} identify domain-relevant experts from a few demonstrations.

\paragraph{MoE upcycling and expansion.} A complementary line of work upcycles dense pretrained models into sparse MoEs. Sparse upcycling initializes an MoE by copying the pretrained feed-forward network of a dense model into multiple experts~\citep{sparseupcycling}, while subsequent work develops improved initialization and training recipes for large-scale upcycling~\citep{upcyclingllm}. Drop-Upcycling partially re-initializes copied experts to break their symmetry and encourage specialization~\citep{dropupcycling}. Other methods construct experts by partitioning dense FFNs rather than replicating them, as in LLaMA-MoE~\citep{llamamoe}. These methods primarily convert dense pretrained checkpoints into MoEs through expert replication, partitioning, and continued training. In contrast, \method starts from an already pretrained MoE and uses target-domain evidence to determine which expert functions should seed the reorganized pool.

\paragraph{Adaptation of pretrained MoEs.} ESFT identifies the experts most relevant to a target domain and updates only those experts while freezing the remaining parameters~\citep{esft2024}. PERFT instead equips pretrained MoEs with routed parameter-efficient adaptation modules~\citep{perft2026}. Both methods adapt a pretrained MoE without changing the composition of its original expert pool. \method reorganizes that pool before standard full-parameter SFT, using domain-relevant experts to initialize the restored expert budget.

\section{Method}
\label{sec:method}

We first set up notation and motivate our approach (\S\ref{sec:overview}), then detail the two operations that reshape the expert pool, namely pruning (\S\ref{sec:prune}) and expansion (\S\ref{sec:expand}), before handing the reorganized model to standard SFT.

\subsection{Preliminaries and Motivation}
\label{sec:overview}

\paragraph{MoE layer.} An MoE layer consists of $E$ routed experts $\{E_1,\dots,E_E\}$ and a router. Given a token representation $h$, the router scores the experts and selects a top-$K$ subset $\mathcal{S}(h)$; the layer output is a gated combination of the selected experts,
\begin{equation}
y \;=\; \sum_{i\in\mathcal{S}(h)} g_i(h)\,E_i(h),
\label{eq:moe}
\end{equation}
where $g_i(h)$ is the normalized gate weight of expert $i$. The expert count $E$ and the active count $K$ are fixed by the pre-trained architecture; together they determine both the parameter count and the per-token inference cost. Any domain adaptation that changes $E$ or $K$ therefore changes the deployment budget, a constraint we deliberately keep.

\paragraph{Motivation.} Pre-training endows an MoE with specialized experts that handle different functions. Under a single target distribution, however, their relevance becomes uneven: some experts exhibit high domain-aligned saliency, whereas others remain poorly aligned despite receiving non-trivial routing. Standard SFT updates this inherited pool in place but does not explicitly reallocate expert slots. This motivates the central idea of \method: reorganize the pool before SFT so that more of its fixed capacity starts from a domain-aligned initialization.

\paragraph{\method in three steps.} Guided by this, \method reorganizes the expert pool into a domain-aligned initialization before SFT, at a fixed budget. \textbf{(1) Prune:} score each expert on a calibration subset $\mathcal{C}\subset\mathcal{D}$ and drop the least domain-salient experts per layer. \textbf{(2) Expand:} regenerate experts from the survivors to restore the original count $E$. \textbf{(3) Domain SFT:} fine-tune the reorganized model on $\mathcal{D}$ with a standard cross-entropy loss. The result is architecturally identical to the original (same $E$, $K$, parameter count, and inference cost), differing only in that its experts are initialized to be domain-aligned rather than generic.
Algorithm~\ref{alg:umoe} in Appendix~\ref{app:algorithm} summarizes the complete procedure.

\subsection{Domain-Aligned Expert Pruning}
\label{sec:prune}

Pruning decides which experts to keep. Because these survivors later rebuild the full pool, we retain those with the highest estimated relevance to the target domain, using a domain-aligned saliency measure rather than routing frequency alone. Following work on expert-level MoE pruning~\citep{reap2025,notallexperts}, we score every expert on a calibration set $\mathcal{C}\subset\mathcal{D}$ and keep the top half per layer.

By default we use the REAP saliency~\citep{reap2025}, which couples an expert's activation strength with its router gate. Let $\mathcal{T}_{l,e}$ be the set of calibration tokens for which expert $e$ is selected at layer $l$, let $f_{l,e}(h_t)$ be its MLP output on token representation $h_t$, and let $g_{l,e}(h_t)$ be its gate weight renormalized over the selected top-$K$ experts. The saliency is the conditional mean
\begin{equation}
S(l,e) \;=\; \frac{1}{|\mathcal{T}_{l,e}|}
\sum_{t\in\mathcal{T}_{l,e}} g_{l,e}(h_t)\,
\bigl\lVert f_{l,e}(h_t)\bigr\rVert_2.
\label{eq:reap}
\end{equation}
We set $S(l,e)=0$ when $\mathcal{T}_{l,e}$ is empty. An expert scores high when it produces a large output under a large router gate when selected. Conditioning on selected tokens separates its per-activation saliency from how frequently it is routed to. We report an ablation over REAP, activation frequency, and unweighted Expert Activation Norm (EAN) in \S\ref{sec:ablation}; REAP and frequency are competitive, while EAN is weaker in our setting.

\subsection{Budget-Preserving Expansion}
\label{sec:expand}

Pruning leaves each layer with $E/2$ domain-salient experts. Expansion regrows the pool back to the original count $E$ so that the model's parameter count, active experts $K$, and inference cost are exactly preserved, with no change to the deployment budget. Regenerating experts from existing weights is reminiscent of upcycling, which initializes MoE experts by copying dense FFN weights~\citep{sparseupcycling,upcyclingllm}, and of partial re-initialization schemes that perturb the copies to promote specialization~\citep{dropupcycling}; the distinction here is that we refill pruned slots with domain-aligned survivors under a fixed budget, turning generic capacity into domain capacity rather than adding capacity.

\paragraph{Regrow, then perturb.} Each freed slot is initialized from a survivor assigned in round-robin order. Under our fixed $50\%$ pruning ratio, each survivor seeds exactly one new expert; its MLP weights and router row are first copied into the freed slot. We then perturb both the parent and the regrown expert independently. For every parent--regrown pair initialized from a parameter tensor $W$, we set
\begin{equation}
W_{\mathrm{parent}}' = W + \sigma\,s(W)\,\boldsymbol{\epsilon}_{p},
\qquad
W_{\mathrm{regrown}}' = W + \sigma\,s(W)\,\boldsymbol{\epsilon}_{r},
\end{equation}
where $\boldsymbol{\epsilon}_{p},\boldsymbol{\epsilon}_{r}\overset{\mathrm{i.i.d.}}{\sim}\mathcal{N}(0,\mathbf{I})$. The scale $s(W)$ is the empirical standard deviation of the corresponding stored parameter group: an individual MLP projection in an unfused implementation, the stored expert tensor in a fused implementation, and the full router matrix for router rows. Noise is sampled independently across parameter groups and between the two sides; we use $\sigma=0.05$. Here, \emph{symmetric} means that the parent and regrown expert are treated symmetrically, not that they receive an antithetic $\pm\boldsymbol{\epsilon}$ pair. For architectures with an expert-score correction bias, the regrown expert also inherits its parent's bias.

\paragraph{Two design principles.} The placement of the perturbation is important to effective expansion. Two observations from the ablations in \S\ref{sec:ablation} motivate our default design:
\begin{itemize}[leftmargin=1.4em,itemsep=2pt,topsep=2pt]
\item \textbf{Perturb both sides, not just the regrown expert.} One-sided perturbation introduces an initialization asymmetry between the unchanged parent and its perturbed counterpart. Empirically, routing to the perturbed half declines during SFT and downstream performance degrades substantially. Independently perturbing both sides removes the privileged unperturbed anchor and treats the pair symmetrically.
\item \textbf{Perturb both the router row and the MLP.} Router and MLP perturbations provide complementary sources of routing and functional asymmetry, respectively. Either component alone performs strongly in our ablation, while their combination gives the best average result by a modest margin.
\end{itemize}

\section{Experiments}
\label{sec:exp}

We first describe the experimental setup in \S\ref{sec:setup}. We then validate \method along three axes: across architectures and domains, and across SFT data scales. Concretely, \S\ref{sec:exp-math} reports the math domain on two MoE architectures (Qwen3-30B-A3B and Qwen3.5-35B-A3B) and a stronger in-house corpus; \S\ref{sec:exp-other} extends to four further domains (code, science, tool-use, and agentic coding) under the same recipe; and \S\ref{sec:exp-scale} studies how the gain varies with SFT data size.

\subsection{Setup}
\label{sec:setup}

\paragraph{Models.} To assess generality across MoE architectures, we use two pre-trained models: Qwen3-30B-A3B~\citep{qwen3} ($E{=}128$ routed experts, $K{=}8$, no shared expert) and Qwen3.5-35B-A3B~\citep{qwen35} ($E{=}256$ routed experts, $K{=}8$, plus a shared expert).

\paragraph{Domains and evaluation.} We evaluate \method on five domains, each with its own SFT dataset and benchmarks. For math, we use a subset of OpenMathReasoning~\citep{openmathreasoning} as the SFT dataset, and evaluate on six competition benchmarks: AIME 2024, AIME 2025, AIME 2026, HMMT Feb 2025, HMMT Nov 2025, and HMMT Feb 2026~\citep{matharena,matharena2026}. For code, we use a subset of OpenCodeReasoning~\citep{opencodereasoning} and evaluate on two LiveCodeBench windows~\citep{livecodebench,livecodebenchrepo}: August 2024 to January 2025 (279 problems, LCB v5) and February to April 2025 (131 problems, LCB v6). For science, we construct a 696,818-example subset of Nemotron-SFT-Science-v2~\citep{nemotronsciencev2} by excluding samples with tool interactions, retaining the first user query and final assistant response, and filtering sequences longer than 32,768 tokens; evaluation uses GPQA-Diamond~\citep{gpqa}. Beyond these reasoning domains, we also evaluate \method on two agentic domains. For tool-use, we use a 119,287-example processed subset of Toucan~\citep{toucan} and evaluate on $\tau$-bench~\citep{taubench} and $\tau^2$-bench~\citep{tau2bench} (airline and retail). For agentic coding, we use the OpenSWE data released with daVinci-Env~\citep{openswe} and evaluate on all 500 instances of SWE-bench Verified~\citep{swebench,swebenchverified} using a locally adapted SWE-agent 1.1.0 scaffold~\citep{sweagent}. Table~\ref{tab:datasets} summarizes the SFT dataset statistics. Unless otherwise noted, evaluations use a sampling temperature of 0.6, top-$p$ of 0.95, and a maximum generation length of 32,768 tokens. The SWE-agent evaluation instead uses a temperature of 0.7, top-$p$ of 1.0, a 131,072-token context window, and at most 100 model calls per instance. The external-model comparison in Table~\ref{tab:math}(b) uses a maximum length of 102,400 tokens. We report avg@32, the mean accuracy over 32 sampled responses per problem, for math; avg@8 analogously for code, science, and tool-use; and pass@1 over all 500 SWE-bench Verified instances.

\begin{table}[h]
\centering\footnotesize
\setlength{\tabcolsep}{5pt}
\begin{tabular}{llrrr}
\toprule
\textbf{Domain} & \textbf{Dataset} & \textbf{Samples} & \textbf{Avg. Length (tokens)} & \textbf{Total Tokens} \\
\midrule
Math & OpenMathReasoning & 300,000 & 6,681 & 2.0B \\
Code & OpenCodeReasoning & 300,000 & 8,705 & 2.6B \\
Science & Nemotron-Science-v2 & 696,818 & 2,743 & 1.9B \\
Tool-use & Toucan & 119,287 & 1,616 & 0.19B \\
Agentic & OpenSWE & 12,358 & 50,355 & 0.62B \\
\bottomrule
\end{tabular}
\caption{SFT dataset statistics for each domain. Token counts are computed with the Qwen3 tokenizer. The same data is used for both REAP calibration (1024-sample subset) and SFT.}
\label{tab:datasets}
\end{table}

\paragraph{Implementation details.} Calibration uses REAP saliency over $1024$ samples (max length 2048) drawn from the corresponding SFT data with a fixed sampling seed of 42. The pruning ratio is $50\%$, and expansion independently perturbs both parents and regrown experts ($\sigma{=}0.05$) in their router rows and MLP weights. For SFT, we train for one epoch with a global batch size of $512$ using the Adam optimizer~\citep{kingma2015adam} ($\beta_1{=}0.9$, $\beta_2{=}0.95$, weight decay $0.1$, gradient clip $0.1$). The learning rate follows a cosine schedule that warms up over the first $10\%$ of steps to a peak of $1\times10^{-4}$ and decays to $1\times10^{-5}$. Training is built on slime~\citep{slime}. The same configuration is applied to all domains with no per-domain tuning; \method and the \directsft baseline use identical SFT data and hyperparameters, differing only in the expert initialization.

\subsection{Main Results}
\label{sec:exp-main}

\subsubsection{Math Domain}
\label{sec:exp-math}

\paragraph{Performance.} Table~\ref{tab:math} reports avg@32 scores---the mean accuracy over 32 sampled responses per problem---on six math competition benchmarks. With a 300k subset of OpenMathReasoning, \method improves over direct SFT on every benchmark and every model, lifting the average accuracy by $+3.40$ on Qwen3-30B-A3B and by $+1.67$ on Qwen3.5-35B-A3B, with a peak improvement of $+6.56$ on AIME 2024. These results demonstrate the effectiveness of our approach across two MoE architectures under the same frozen recipe.

Beyond open-source data, we further evaluate on a high-quality, large-scale in-house math corpus ($\sim$150k samples, average response length $\sim$40k tokens). Checkpoints trained on this data are already competitive with strong open-source models at a similar parameter scale: as shown in Table~\ref{tab:math}(b), our \directsft baseline has a higher six-benchmark average than Qwen3-30B-A3B-Thinking (82.81 vs.\ 81.06). In this strong setting, \method improves five of the six benchmarks and lifts the math-benchmark average by $+1.36$, with the largest single-benchmark improvement of $+3.03$ on HMMT Feb 2026. This confirms that expert-pool reorganization remains effective when paired with high-quality SFT data and a strong baseline.

\begin{table}[h]
\centering\small
\resizebox{\columnwidth}{!}{%
\setlength{\tabcolsep}{3.5pt}
\begin{tabular}{llccccccc}
\toprule
\textbf{Model} & \textbf{Method} & \textbf{AIME24} & \textbf{AIME25} & \textbf{AIME26} & \textbf{HMMT-F25} & \textbf{HMMT-N25} & \textbf{HMMT-F26} & \textbf{Avg.} \\
\midrule
\multicolumn{9}{c}{\textit{(a) OpenMathReasoning (300k subset)}} \\[2pt]
\multirow{2}{*}{Qwen3-30B-A3B} & \directsft & 58.13 & 40.94 & 50.83 & 39.27 & 38.65 & 30.30 & 43.02 \\
 & \method & \textbf{64.69} & \textbf{47.08} & \textbf{53.44} & \textbf{41.35} & \textbf{40.52} & \textbf{31.44} & \textbf{46.42} \\
 & $\Delta$ & \textcolor{gray}{+6.56} & \textcolor{gray}{+6.14} & \textcolor{gray}{+2.61} & \textcolor{gray}{+2.08} & \textcolor{gray}{+1.87} & \textcolor{gray}{+1.14} & \textcolor{gray}{+3.40} \\
\arrayrulecolor{black!25}\cmidrule{1-9}\arrayrulecolor{black}
\multirow{2}{*}{Qwen3.5-35B-A3B} & \directsft & 78.65 & 70.99 & 71.20 & 56.35 & 54.69 & 38.64 & 61.75 \\
 & \method & \textbf{80.31} & \textbf{72.03} & \textbf{71.72} & \textbf{58.23} & \textbf{58.75} & \textbf{39.49} & \textbf{63.42} \\
 & $\Delta$ & \textcolor{gray}{+1.66} & \textcolor{gray}{+1.04} & \textcolor{gray}{+0.52} & \textcolor{gray}{+1.88} & \textcolor{gray}{+4.06} & \textcolor{gray}{+0.85} & \textcolor{gray}{+1.67} \\
\midrule
\multicolumn{9}{c}{\textit{(b) Inhouse Math Data}} \\[2pt]
Qwen3-30B-A3B-Thinking~\citep{qwen3} & & 91.35 & 86.04 & 86.98 & 68.96 & 80.42 & 72.63 & 81.06 \\
SU-01~\citep{su012026} & & 93.54 & 91.88 & 90.94 & 91.67 & 88.23 & 77.65 & 88.99 \\
Nex-N2-Mini~\citep{nexn2mini2026} & & 86.25 & 81.98 & 80.94 & 77.08 & 75.52 & 76.04 & 79.64 \\
Qwen3.6-35B-A3B~\citep{qwen36} & & 94.90 & 91.56 & 91.88 & 88.96 & 85.94 & 81.53 & 89.13 \\
\arrayrulecolor{black!25}\cmidrule{1-9}\arrayrulecolor{black}
\multirow{2}{*}{Qwen3-30B-A3B} & \directsft & 91.56 & 86.88 & 85.31 & \textbf{81.04} & 79.17 & 72.92 & 82.81 \\
 & \method & \textbf{92.60} & \textbf{88.33} & \textbf{87.92} & 80.00 & \textbf{80.21} & \textbf{75.95} & \textbf{84.17} \\
 & $\Delta$ & \textcolor{gray}{+1.04} & \textcolor{gray}{+1.45} & \textcolor{gray}{+2.61} & \textcolor{gray}{$-$1.04} & \textcolor{gray}{+1.04} & \textcolor{gray}{+3.03} & \textcolor{gray}{+1.36} \\
\bottomrule
\end{tabular}%
}
\caption{Avg@32 scores on six math competition benchmarks. All models in block (b), including open-source checkpoints, are evaluated with a maximum generation length of 102,400. Boldface within the final two rows compares \directsft and \method.}
\label{tab:math}
\end{table}

\subsubsection{Cross-Domain Validation}
\label{sec:exp-other}

\begin{table}[h]
\centering\footnotesize
\setlength{\tabcolsep}{5pt}
\begin{tabular}{llccccccc}
\toprule
 & & \multicolumn{2}{c}{\textbf{Code}} & \textbf{Science} & \multicolumn{2}{c}{\textbf{Tool-use}} & \textbf{Agentic} & \\
\cmidrule(lr){3-4}\cmidrule(lr){5-5}\cmidrule(lr){6-7}\cmidrule(lr){8-8}
\textbf{Model} & \textbf{Method} & LCB v5 & LCB v6 & GPQA-D & $\tau$-bench & $\tau^2$-bench & SWE-Verified & \makecell{\textbf{Unweighted}\\\textbf{Avg.}} \\
\midrule
\multirow{2}{*}{Qwen3-30B-A3B} & \directsft & 49.55 & 44.28 & 61.36 & 14.31 & 25.33 & 16.2 & 35.17 \\
 & \method & \textbf{52.28} & \textbf{45.13} & \textbf{64.46} & \textbf{15.51} & \textbf{26.73} & \textbf{22.2} & \textbf{37.72} \\
\arrayrulecolor{black!25}\cmidrule{1-9}\arrayrulecolor{black}
\multicolumn{2}{l}{$\Delta$} & \textcolor{gray}{+2.73} & \textcolor{gray}{+0.85} & \textcolor{gray}{+3.10} & \textcolor{gray}{+1.20} & \textcolor{gray}{+1.40} & \textcolor{gray}{+6.0} & \textcolor{gray}{+2.55} \\
\bottomrule
\end{tabular}
\caption{Cross-domain results on Qwen3-30B-A3B. Code: LCB v5 (August 2024 to January 2025) and v6 (February to April 2025), both with avg@8. Science: GPQA-Diamond (avg@8). Tool-use: $\tau$-bench/$\tau^2$-bench airline+retail average (avg@8). Agentic: SWE-bench Verified pass@1 over 500 instances. The final column is the unweighted average over the six benchmark columns despite their different evaluation metrics.}
\label{tab:other}
\end{table}

We further validate \method on Qwen3-30B-A3B across additional domains, covering both reasoning tasks (code, science) and agentic tasks (tool-use, agentic coding), using the same frozen recipe with only the calibration and SFT data changed per domain. As shown in Table~\ref{tab:other}, \method is positive on every evaluated benchmark: code $+2.73$/$+0.85$ on LiveCodeBench v5/v6, science $+3.10$ on GPQA-Diamond, tool-use $+1.20$/$+1.40$ on $\tau$-bench/$\tau^2$-bench (averaged over airline and retail), and agentic coding improves by 6.0 points in pass@1 on SWE-bench Verified (16.2\%$\to$22.2\%). These results suggest that expert-pool reorganization is not specific to math and can transfer across diverse reasoning and agentic domains.

\subsubsection{Data Scaling}
\label{sec:exp-scale}

\begin{figure}[h]
\centering
\includegraphics[width=\columnwidth]{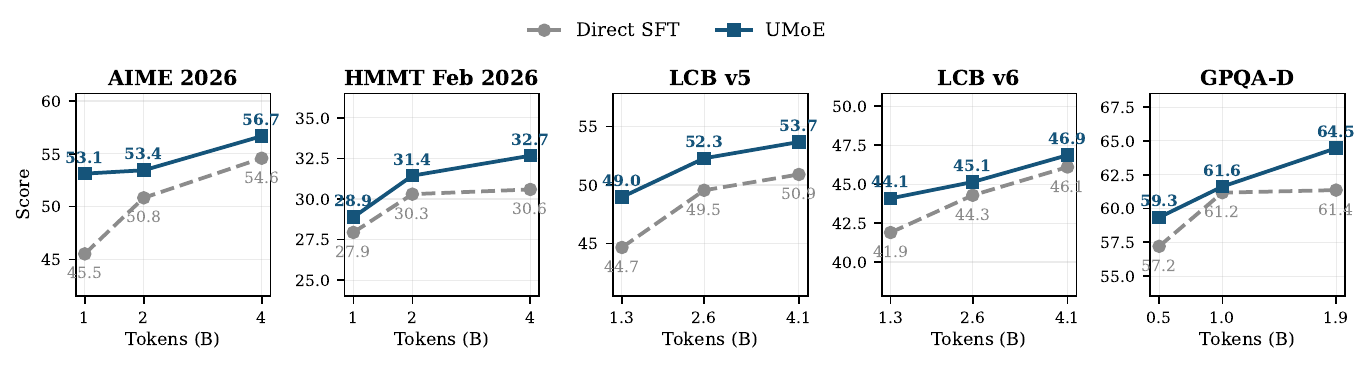}
\caption{Data scaling across three domains and five benchmarks (Qwen3-30B-A3B). Each panel shows \directsft (gray dashed) vs.\ \method (blue solid) at increasing SFT data sizes (horizontal axis: billion tokens). All 15 paired comparisons favor \method.}
\label{fig:scaling}
\end{figure}

To examine whether \method's benefit persists across data regimes, we evaluate it on three domains (math, code, and science) at multiple SFT data scales, ranging from 0.5B to 4.1B tokens depending on the domain (Figure~\ref{fig:scaling}). All 15 paired comparisons favor \method. For AIME 2026, the improvement is $+7.60$ at 1B tokens and remains $+2.09$ at 4B; for LCB v5, it is $+4.30$ at 1.3B and $+2.77$ at 4.1B; for GPQA-Diamond, it is $+2.14$ at 0.5B and $+3.10$ at 1.9B. Thus, the gain remains positive throughout the evaluated data range.

\section{Analysis}
\label{sec:analysis}

We analyze \method along three axes. \S\ref{sec:ablation} ablates the two design choices: the pruning criterion and how to expand. \S\ref{sec:functional} examines whether the gains are associated with more effective use of the fixed expert pool. \S\ref{sec:behavior} asks how the gain appears in the model's outputs.

\subsection{Ablation Study}
\label{sec:ablation}

All ablations are conducted on Qwen3-30B-A3B, fine-tuned on OpenMathReasoning and evaluated on the math benchmarks, with all remaining SFT hyperparameters held fixed. We ablate the two algorithmic steps independently: when varying the pruning criterion we fix expansion to symmetric $\sigma{=}0.05$ noise on router and MLP rows; when varying the expansion we fix pruning to REAP-50. The default configuration (REAP pruning with symmetric router$+$MLP noise) is the shared anchor row in Table~\ref{tab:ablation}, and every variant is measured against \directsft.

\begin{table}[h]
\centering\small
\begin{tabular}{llrr}
\toprule
\textbf{Ablation axis} & \textbf{Variant} & \textbf{AVG6} & \textbf{$\Delta$} \\
\midrule
\multicolumn{2}{l}{\directsft} & 37.85 & $-$ \\
\midrule
\multirow{3}{*}{Pruning criterion}
 & EAN mean ($\mathbb{E}[\|f_e\|\mid e\text{ active}]$) & 40.39 & +2.54 \\
 & Frequency ($a_e$)                   & 42.63 & +4.78 \\
 & REAP ($\mathbb{E}[g_e\|f_e\|\mid e\text{ active}]$, default) & \textbf{43.01} & \textbf{+5.16} \\
\midrule
\multirow{4}{*}{Expansion}
 & Regrown-side noise only    & 33.61 & $-4.24$ \\
 & Router only (sym.)         & 42.78 & +4.93 \\
 & MLP only (sym.)            & 42.71 & +4.86 \\
 & Router\,$+$\,MLP (default, sym.) & \textbf{43.01} & \textbf{+5.16} \\
\bottomrule
\end{tabular}
\caption{Ablation of the two algorithmic steps (Qwen3-30B-A3B, math OpenMath 150k, math-benchmark average at $n{=}32$). Pruning-criterion rows fix expansion to symmetric $\sigma{=}0.05$; expansion-noise rows fix pruning to REAP-50. The default (REAP $+$ router$+$MLP) is the shared anchor.}
\label{tab:ablation}
\end{table}

\paragraph{Pruning criterion.}
The prune step ranks experts by a domain saliency score, and we compare three criteria. Activation frequency ($a_e$) counts how often the router selects expert $e$ on the calibration set. EAN mean instead measures the conditional mean $\ell_2$ norm of the expert's output over tokens on which it is selected, capturing activation strength independently of usage frequency. REAP computes the conditional mean of this output norm weighted by the corresponding router gate. All three improve over direct SFT (Table~\ref{tab:ablation}), but not equally. Plain frequency is already a strong signal ($+4.78$), showing that how often the router uses an expert is a useful proxy for domain relevance. EAN mean is weaker ($+2.54$), whereas adding the router gate in REAP gives the best result ($+5.16$), marginally ahead of frequency; we therefore adopt REAP as the default.

\begin{figure}[t]
\centering
\includegraphics[width=\columnwidth]{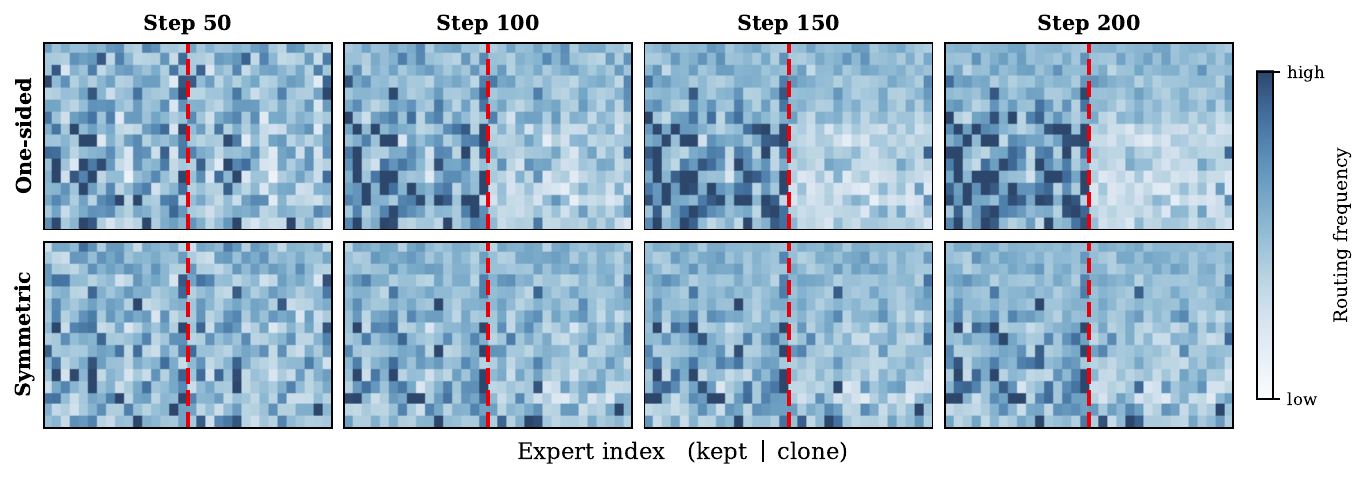}
\caption{Per-(layer, expert) routing frequency over SFT (downsampled; the red line marks the parent $\mid$ regrown boundary). Under one-sided noise (top), routing to the regrown half progressively declines; symmetric noise (bottom) maintains a more balanced allocation between the two halves.}
\label{fig:onesided_drift}
\end{figure}

\paragraph{Expansion.}
Expansion restores the pruned pool to its original size by initializing one regrown expert from each retained parent. We study two aspects of the perturbation: which parameters receive noise (the router row, the MLP weights, or both), and whether noise is applied only to the regrown side (one-sided) or independently to both sides (symmetric). Table~\ref{tab:ablation} reports all variants. Among symmetric variants, perturbing both the router and the MLP is best ($43.01$), while router-only ($42.78$) and MLP-only ($42.71$) are close behind. This is consistent with the two noise sites providing complementary routing and functional asymmetry, although the margin from combining them is modest. One-sided noise, in contrast, scores only $33.61$, below direct SFT. Figure~\ref{fig:onesided_drift} shows the associated routing dynamics: when only the regrown side is perturbed, its routing share progressively declines during SFT. Independently perturbing both sides removes this initial asymmetry and maintains a more balanced routing allocation.

\subsection{Mechanism Analysis}
\label{sec:functional}

\method leaves the number of experts, the top-$K$, and the compute budget unchanged. We therefore examine whether its gains are associated with more effective use of the fixed expert pool. Our results provide evidence that \directsft allocates substantial routed-expert compute to a redundant subset, whereas a larger fraction of \method's pool becomes consequential to downstream performance.

\begin{figure}[t]
\centering
\includegraphics[width=\columnwidth]{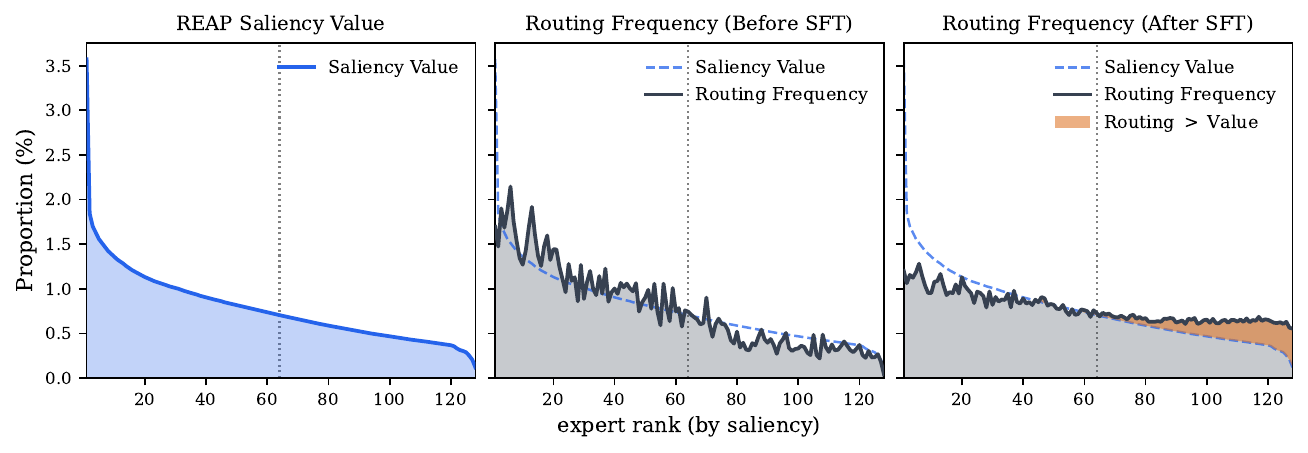}
\caption{The direct-SFT model allocates routing to a low-saliency subset. Experts are ranked by REAP saliency (left). Before SFT (middle), routing frequency is concentrated on the salient experts. After SFT (right), routing becomes flatter and the low-saliency tail receives a larger share of routing than of normalized saliency (orange). The dashed line repeats the saliency profile for reference; the dotted line marks the top-64 experts kept by pruning.}
\label{fig:waste}
\end{figure}

\paragraph{The direct-SFT pool contains redundant routed capacity.}
\method removes the half of the expert pool with the lowest target-domain REAP saliency at initialization, so we track exactly this subset through math SFT. Let $\mathcal{B}_l$ denote this bottom-half set in layer $l$. For a per-expert metric $m_{l,e}$ (routing count or REAP saliency), we report the layer-normalized share $\frac{1}{L}\sum_{l=1}^{L}\frac{\sum_{e\in\mathcal{B}_l}m_{l,e}}{\sum_{e=1}^{E}m_{l,e}}$, with $L{=}48$. Under this aggregation, the subset initially receives $25.7\%$ of the math routing. During SFT its routing share climbs to $41.6\%$ (Figure~\ref{fig:waste}), while it accounts for $33.0\%$ of the REAP saliency after SFT. Thus, its share of routed-expert assignments is substantially larger than its share of the measured saliency. Because expert-FFN cost scales with routing, approximately $42\%$ of the routed-expert computations are allocated to this subset. The post-hoc pruning experiment below tests whether that allocation is functionally redundant.

\paragraph{Post-hoc pruning tests redundancy and locates the gain.}
To test whether this half is truly redundant, we take the final SFT checkpoints and prune each one back to its own top-64 experts by REAP saliency, then re-evaluate on the math benchmarks (Table~\ref{tab:knockout}).

\begin{table}[h]
\centering\small
\resizebox{\columnwidth}{!}{%
\setlength{\tabcolsep}{3.5pt}
\begin{tabular}{llccccccc}
\toprule
\textbf{Model} & \textbf{Experts} & \textbf{AIME24} & \textbf{AIME25} & \textbf{AIME26} & \textbf{HMMT-F25} & \textbf{HMMT-N25} & \textbf{HMMT-F26} & \textbf{Avg.} \\
\midrule
\multirow{2}{*}{\directsft} & 128 & 51.46 & 38.54 & 45.52 & 32.29 & 31.35 & 27.94 & 37.85 \\
 & 64 (pruned) & 51.77 & 39.27 & 45.83 & 30.73 & 32.19 & 26.61 & 37.73 \\
\multicolumn{2}{l}{\textcolor{gray}{$\Delta$}} & \textcolor{gray}{+0.31} & \textcolor{gray}{+0.73} & \textcolor{gray}{+0.31} & \textcolor{gray}{$-$1.56} & \textcolor{gray}{+0.84} & \textcolor{gray}{$-$1.33} & \textcolor{gray}{$-$0.12} \\
\midrule
\multirow{2}{*}{\method} & 128 & \textbf{60.00} & \textbf{42.19} & \textbf{53.12} & \textbf{36.04} & \textbf{37.81} & \textbf{28.88} & \textbf{43.01} \\
 & 64 (pruned) & 55.10 & 39.06 & 48.75 & 27.81 & 30.52 & 26.80 & 38.01 \\
\multicolumn{2}{l}{\textcolor{gray}{$\Delta$}} & \textcolor{gray}{$-$4.90} & \textcolor{gray}{$-$3.13} & \textcolor{gray}{$-$4.37} & \textcolor{gray}{$-$8.23} & \textcolor{gray}{$-$7.29} & \textcolor{gray}{$-$2.08} & \textcolor{gray}{$-$5.00} \\
\bottomrule
\end{tabular}%
}
\caption{Post-SFT pruning: each model is pruned to its own top-64 experts by REAP saliency (removing the least-salient half), then re-evaluated (Qwen3-30B-A3B, math benchmarks, avg@32). Pruning \directsft changes the average by only $-0.12$, whereas pruning \method costs 5.00 points, collapsing the gap from $+5.16$ to $+0.28$.}
\label{tab:knockout}
\end{table}

Pruning \directsft to 64 experts changes its six-benchmark average by only $-0.12$, despite removing the subset that received roughly $40\%$ of its routing. Together with the routing--saliency mismatch above, this indicates substantial redundancy in how \directsft uses its expert pool. Pruning \method, in contrast, costs $5.0$ points, and the gap over \directsft collapses from $+5.16$ (128 experts) to $+0.28$ (64 experts). This supports the interpretation that most of \method's advantage depends on making the additional half of its expert pool collectively consequential, rather than merely improving a core set that remains equally strong after half the pool is removed.

\begin{figure}[t]
\centering
\begin{minipage}[t]{0.48\columnwidth}
\centering
\includegraphics[width=\linewidth]{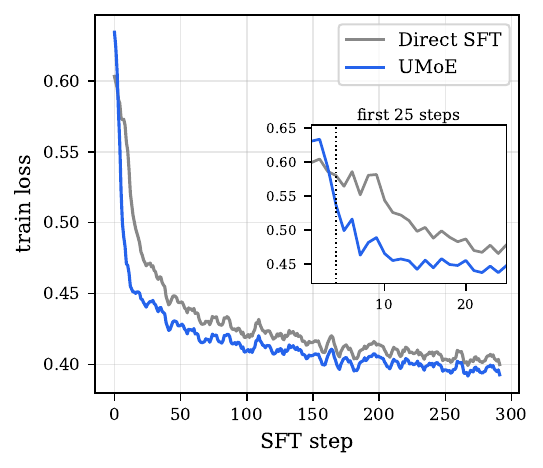}
\caption{Math SFT loss (150k): \method starts higher, then trains below \directsft and ends lower (0.396 vs.\ 0.404).}
\label{fig:loss}
\end{minipage}\hfill
\begin{minipage}[t]{0.48\columnwidth}
\centering
\includegraphics[width=\linewidth]{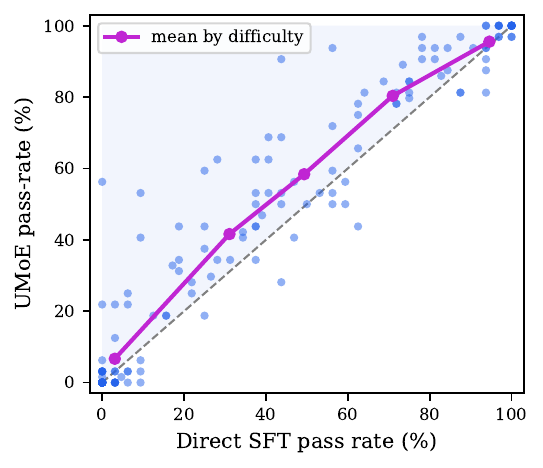}
\caption{Per-problem pass rate. When problems are grouped by their observed \directsft pass rate, the per-bin mean stays above the diagonal in every bin.}
\label{fig:reliability}
\end{minipage}
\end{figure}

\paragraph{Optimization dynamics.}
The reorganized pool follows a different optimization trajectory. As Figure~\ref{fig:loss} shows, \method starts at a higher loss than \directsft in the first few steps, since reorganizing the pool without any training slightly degrades the model. Within a few steps, however, its loss drops rapidly, falls below \directsft, and stays lower through the rest of the run, reaching a slightly lower final training loss (0.396 vs.\ 0.404).


\subsection{Behavioral Analysis}
\label{sec:behavior}

We analyze the two models' behavior on the math benchmarks to pin down where \method's gains come from (\directsft vs.\ \method).

\paragraph{Gains across observed \directsft pass rates.}
Figure~\ref{fig:reliability} plots every problem's pass rate under the two models: each point is one problem, with its observed \directsft pass rate on the x-axis and its \method pass rate on the y-axis; points above the diagonal favor \method. To summarize the descriptive trend, we bin problems into five equal-width intervals of \directsft pass rate and plot the per-bin mean. The mean stays above the diagonal in every bin, with the largest gains of 9 to 11 points in the middle bins.

\definecolor{casered}{HTML}{B08A8A}
\definecolor{casegreen}{HTML}{8AA890}
\definecolor{errred}{HTML}{C0392B}
\definecolor{okgreen}{HTML}{2E7D32}

\paragraph{An illustrative example.}
The following high-gain problem, AIME 2026 Problem 25~\citep{matharena2026}, illustrates how a difference can arise from one pivotal move. In the displayed solutions, the derivations agree up to a ratio from the Angle Bisector Theorem: \directsft reduces it incorrectly (marked in \textcolor{errred}{red}) and propagates the error to a wrong answer, whereas \method performs the reduction correctly.

\begin{tcolorbox}[colback=black!3, colframe=black!45, boxrule=0.6pt, arc=1pt, left=5pt, right=5pt, top=3pt, bottom=3pt, title=\textbf{Problem} \textnormal{(AIME 2026, Problem 25; correct answer $850$)}, fonttitle=\small, coltitle=black]
\small In $\triangle ABC$, $D$ lies on $\overline{BC}$ with $\overline{AD}$ bisecting $\angle BAC$. A circle $\omega$ passes through $A$ and is tangent to $\overline{BC}$ at $D$, meeting $\overline{AB}$ again at $E$ and $\overline{AC}$ again at $F$. Given $AB=200$, $AC=225$, and $AE,AF,BD,CD$ all positive integers, find the sum of all possible values of $BC$.
\end{tcolorbox}

\vspace{2pt}
\begin{center}
\begin{minipage}[t]{0.478\columnwidth}
\begin{tcolorbox}[colback=casered!6, colframe=casered, title=\textcolor{errred}{\ding{55}}~\textbf{Direct SFT} \textnormal{(before)}~$\to~\textcolor{errred}{405}$, fonttitle=\small, boxrule=0.8pt, arc=1pt, left=4pt, right=4pt, top=3pt, bottom=3pt]
\small By the Angle Bisector Theorem, $\dfrac{BD}{DC}=\dfrac{AB}{AC}=\dfrac{200}{225}$, which \directsft reduces to
\[ \frac{200}{225}=\textcolor{errred}{\mathbf{\tfrac{4}{5}}}\quad\textcolor{errred}{\text{(wrong)}}. \]
Then $BD=\tfrac{4}{9}BC$, $DC=\tfrac{5}{9}BC$, so $BC$ is forced to be a multiple of $9$. This wrong divisibility propagates through the count and the run settles on $\textcolor{errred}{405}$ ($12/32$ samples).
\end{tcolorbox}
\end{minipage}\hfill
\begin{minipage}[t]{0.478\columnwidth}
\begin{tcolorbox}[colback=casegreen!6, colframe=casegreen, title=\textcolor{okgreen}{\ding{51}}~\textbf{\method} \textnormal{(after)}~$\to~\textcolor{okgreen}{850}$, fonttitle=\small, boxrule=0.8pt, arc=1pt, left=4pt, right=4pt, top=3pt, bottom=3pt]
\small By the Angle Bisector Theorem, $\dfrac{BD}{DC}=\dfrac{AB}{AC}=\dfrac{200}{225}$, which \method reduces to
\[ \frac{200}{225}=\textcolor{okgreen}{\mathbf{\tfrac{8}{9}}}\quad\textcolor{okgreen}{\text{(correct)}}. \]
Then $BD=\tfrac{8}{17}BC$, $DC=\tfrac{9}{17}BC$, so $BC$ is a multiple of $17$. With the correct ratio the divisibility follows and the run reaches $\textcolor{okgreen}{850}$ ($29/32$ samples).
\end{tcolorbox}
\end{minipage}
\end{center}

\section{Conclusion}

We presented \method, a budget-preserving approach that reorganizes the expert pool of a pretrained MoE before domain SFT. It prunes the least domain-salient half of the routed experts, regrows the pool from the retained experts with independent perturbations applied to each parent and regrown expert, and then performs standard full-parameter SFT. The resulting model preserves the original architecture, parameter count, routed expert count, top-$K$, and inference cost. We use the same reorganization recipe across all evaluated settings without domain-specific hyperparameter tuning.

Across two MoE architectures, five capability domains, and 12 benchmarks, \method consistently improves over \directsft. Representative improvements include 3.4 points in math average accuracy, 6.0 points on SWE-bench Verified, and 1.67 points on Qwen3.5-35B-A3B. The gains remain positive across all evaluated data scales and on a strong in-house math corpus where \directsft already surpasses Qwen3-30B-A3B-Thinking. Our analysis attributes these improvements to more effective use of the fixed expert budget: \directsft assigns substantial routing to experts that can be removed with little effect, whereas pruning half of \method's trained expert pool causes a 5.0-point drop. These results suggest that reorganizing inherited expert capacity can improve domain adaptation without increasing inference cost. Exploring adaptive pruning ratios and architecture-specific perturbation strategies remains promising future work.

\bibliographystyle{acl_natbib}
\bibliography{references}

\clearpage
\appendix

\section{Full Algorithm}
\label{app:algorithm}

Algorithm~\ref{alg:umoe} consolidates the pruning, regrowth, and fine-tuning operations described in \S\ref{sec:method}. It introduces no auxiliary training objective and restores the original expert count before SFT.

\begin{algorithm}[H]
\caption{Budget-preserving domain specialization with \method.}
\label{alg:umoe}
\small
\textbf{Input:} Pre-trained MoE model $M$ with $L$ layers, $E$ routed experts per layer, and top-$K$ routing; target-domain SFT data $\mathcal{D}$; calibration subset $\mathcal{C}\subset\mathcal{D}$; perturbation scale $\sigma$.\\
\textbf{Output:} Domain-specialized model $M^{\star}$ with the same architecture and inference budget as $M$.
\begin{enumerate}[leftmargin=1.5em,itemsep=2pt,topsep=3pt]
    \item Run $M$ on $\mathcal{C}$. For every layer $l$ and routed expert $e$, collect the selected-token set $\mathcal{T}_{l,e}$, the top-$K$-normalized gate weights, and the expert-output norms.
    \item Compute the domain saliency $S(l,e)$ using Eq.~\ref{eq:reap}; set $S(l,e)=0$ if $\mathcal{T}_{l,e}$ is empty.
    \item For each layer $l$, retain the $E/2$ experts with the largest saliency values and remove the remaining $E/2$ experts.
    \item Assign each retained expert to one freed slot. Under the $50\%$ pruning ratio, every retained expert seeds exactly one regrown expert.
    \item For every parent--regrown pair and parameter group $W$, initialize both sides from the retained expert and independently perturb them:
    \[
    W_{p}'=W+\sigma s(W)\epsilon_{p},\qquad
    W_{r}'=W+\sigma s(W)\epsilon_{r},
    \quad \epsilon_{p},\epsilon_{r}\overset{\mathrm{i.i.d.}}{\sim}\mathcal{N}(0,I).
    \]
    Apply the same procedure to the corresponding router rows, restoring $E$ routed experts in every layer.
    \item Fine-tune all parameters of the reorganized model on $\mathcal{D}$ using the standard next-token cross-entropy objective, producing $M^{\star}$.
\end{enumerate}
\textbf{Calibration configuration:} $|\mathcal{C}|=1024$, maximum calibration length $2048$, sampling seed $42$, pruning ratio $50\%$, and $\sigma=0.05$.
\end{algorithm}

\end{document}